\pdfoutput=1

\documentclass[11pt]{article}

\usepackage[]{ACL2023}

\usepackage{float}
\usepackage{color}
\usepackage{amssymb}
\usepackage{natbib}
\usepackage{booktabs}
\usepackage{tabularx}
\usepackage{amsmath}
\usepackage{multirow}
\usepackage{url}
\usepackage{times}
\usepackage{latexsym}
\usepackage{graphicx}
\usepackage[T1]{fontenc}

\usepackage[utf8]{inputenc}

\usepackage{microtype}

\usepackage{inconsolata}

\title{Emotion Recognition in Signers}

\author{
  Kotaro Funakoshi\\
  FIRST, Institute of Integrated Research\\
  Institute of Science Tokyo \\
  \texttt{funakoshi@first.iir.isct.ac.jp} \\\And
  Yaoxiong Zhu\\
  ICT, School of Engineering\\
  Institute of Science Tokyo \\
  \texttt{\ \ \ zhuyaoxiong@lr.first.isct.ac.jp} \\
}

\usepackage{comment}

\begin{document}
\maketitle
\begin{abstract}
Recognition of signers' emotions suffers from one theoretical challenge and one practical challenge, namely, the overlap between grammatical and affective facial expressions and the scarcity of data for model training. This paper addresses these two challenges in a cross-lingual setting using our eJSL dataset, a new benchmark dataset for emotion recognition in Japanese Sign Language signers, and BOBSL, a large British Sign Language dataset with subtitles. In eJSL, two signers expressed 78 distinct utterances with each of seven different emotional states, resulting in 1,092 video clips. We empirically demonstrate that 1) textual emotion recognition in spoken language mitigates data scarcity in sign language, 2) temporal segment selection has a significant impact, and 3) incorporating hand motion enhances emotion recognition in signers. Finally we establish a stronger baseline than spoken language LLMs (Qwen 2.5 and GPT-4o). 
\end{abstract}

\section{Introduction}

Emotion recognition
is a core topic not only in natural language processing~\citep{yun2024telme} but also in affective computing and human-computer interaction~\citep{zeng2009survey, el2011survey}, enabling more natural and empathetic systems. 
Such systems are equally or more important for social minorities.
Recently, more light is shed on sign language~\citep{long2024skeletonface,Yin2024,Zhen2025}, however, automatic emotion recognition in signers has not been explored at all.
To our best knowledge, the single contribution in this direction is the EmoSign dataset for American Sign Language (ASL)~\cite{chua2025emosign}.

In this paper, we introduce eJSL\footnote{
\url{https://dataverse.harvard.edu/dataverse/eJSL}
}, 
a new benchmark dataset for emotion recognition in Japanese Sign Language JSL).
We asked two signers to express 78 distinct sentences with each of seven different emotional states, resulting in 1,092 video clips. 
Because human languages are highly context-dependent, any linguistic expression potentially can be expressed with any emotion.
In this dataset, thus, the task is 
recognizing the emotions of signing signers rather than that of signed contents.

Here, the arising challenge is that
emotion expressions in signers
are further complicated because facial expressions convey both grammatical and affective information~\citep{brentari2009prosodic, wilbur2000phonological}. For example, eyebrow movement can signal a yes/no question~\citep{Pfau_Quer_2010} or express surprise~\citep{valli2000linguistics}, creating ambiguity for 
emotion recognition models trained on non-signers.

To address the challenge, we investigate three  hypotheses: (1) caption-based weakly labeled data can support effective model fine-tuning, (2) selecting temporal segments less affected by grammatical expressions improve accuracy, and (3) hand gesture features enhance recognition beyond facial features alone. Experiments on multiple datasets
validate these hypotheses and offer insights into understanding 
of emotional communication in signers.

\section{Emotion Recognition and Sign Language}
\label{sec:er_and_sl}
As 
discussed,
a unique challenge in emotion recognition in signers lies in the overlap between grammatical facial expressions (GFEs) and affective facial expressions (AFEs). Unlike spoken language, sign language uses non-manual markers such as facial movements and head gestures to encode syntax. 
These signals often occur simultaneously with AFEs, 
making their separation critical for accurate understanding of communicative information.  

To this end, \citet{silva2020recognition}
annotated their corpus 
with facial Action Units (AUs) to encode GFEs.
However, this corpus is not annotated in terms of emotion.
Although there are many other sign language datasets (see Table 2 of \cite{albanie-etal-2021-bobsl} for a not-exaustive but rich list of 30 datasets), none of them are with emotion annotation except for EmoSign and our eJSL. However, both of them are small-scale benchmark-oriented datasets.
Thus the scarcity of data available for supervised training is another challenge. 

In human multimodal communication, verbal and non-verbal information can be independent, even contradictory in sentiment. 
In such contradictory situations, facial information can be highly dominant more than verbal information~\citep{mehrabian1971silent}.
Nevertheless, in usual situations, it is repeatedly observed that textual information is dominant by the multimodal spoken language emotion recognition literature~\citep{Li2023,yun2024telme}.
Therefore, we can expect that, even in sign language, 
textual caption/subtitle data (i.e., translations in spoken language) are useful to induce the emotion state of original signers in existing corpora.
In this paper, we explore this possibility.

\begin{figure*}[t]
\centering
\resizebox{0.9\linewidth}{!}{
\setlength{\tabcolsep}{3pt} 
\newcommand{\figw}{0.20\textwidth}
\bf
\begin{tabular}{cccccc}
  Joy &   
  Sadness &   
  Anger &   
  Disgust & 
  Fear & 
  Surprise  
  \\
  \includegraphics[width=\figw,trim={4cm 0 4cm 3cm},clip]{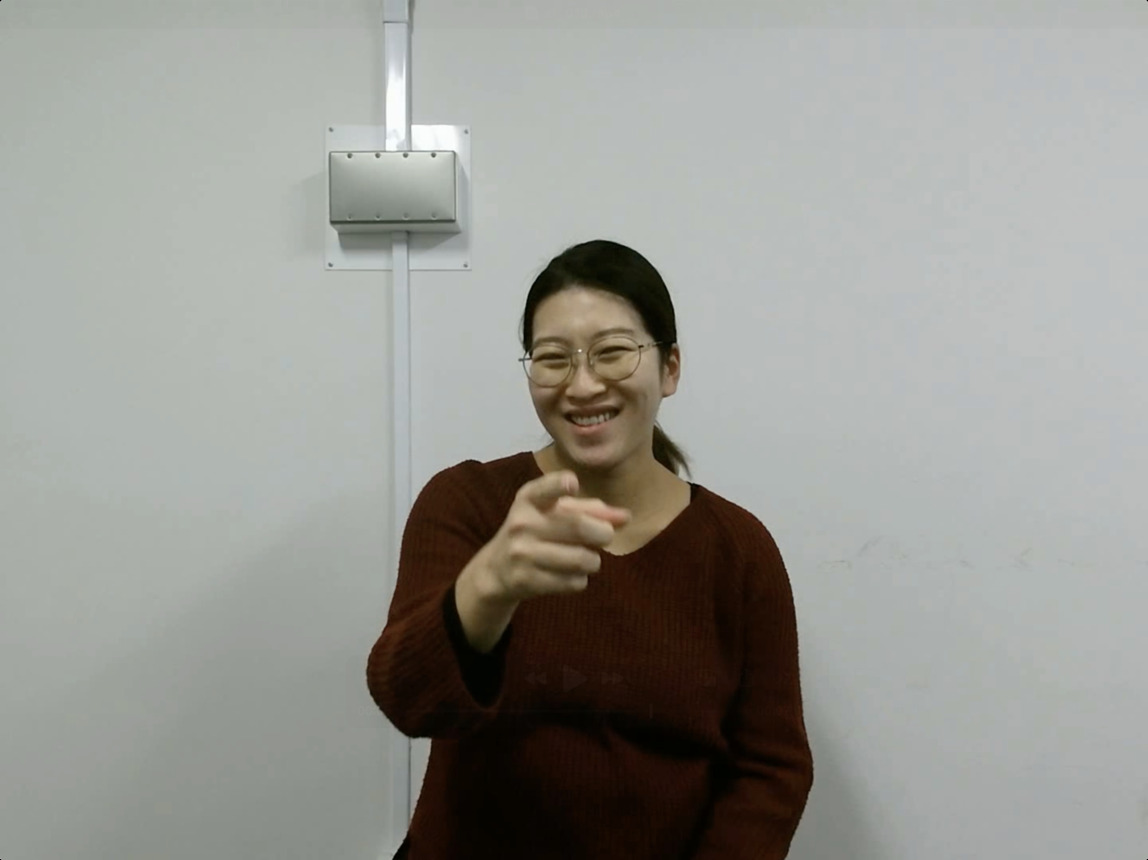} &
  \includegraphics[width=\figw,trim={4cm 0 4cm 3cm},clip]{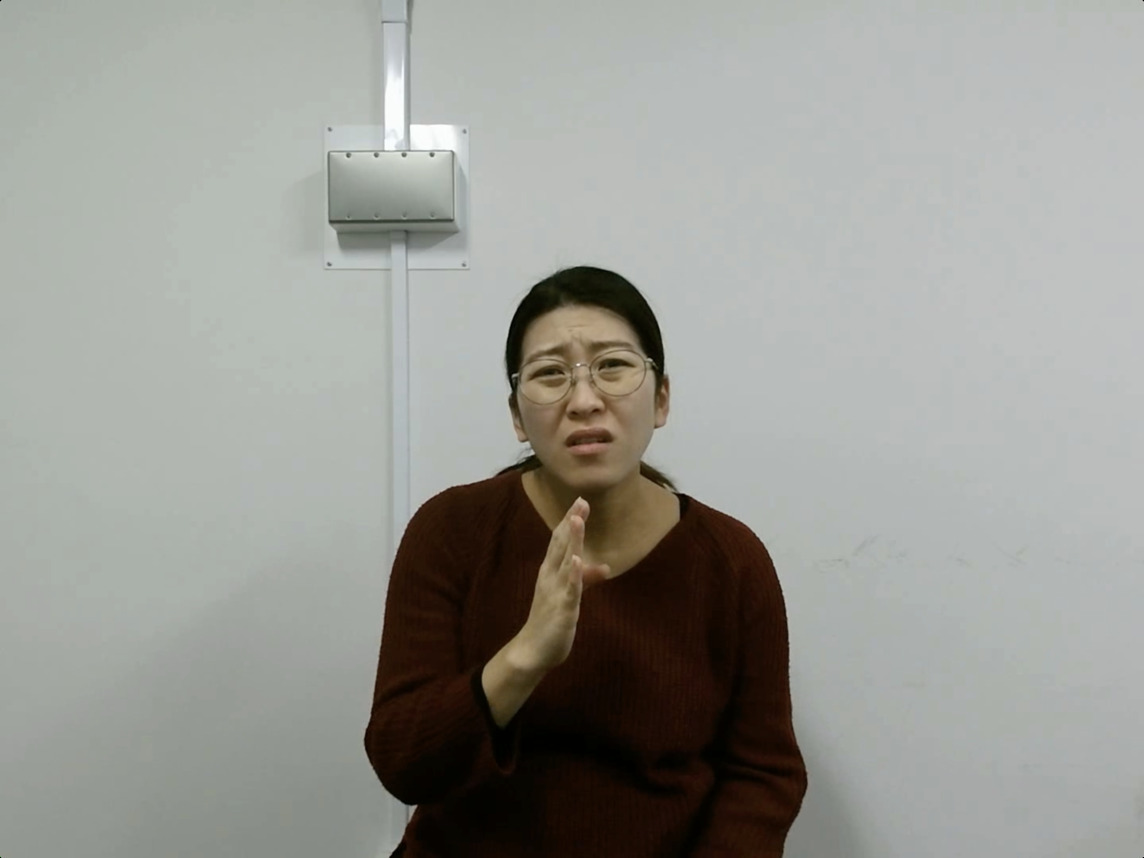} &
  \includegraphics[width=\figw,trim={8cm 0 8cm 6cm},clip]{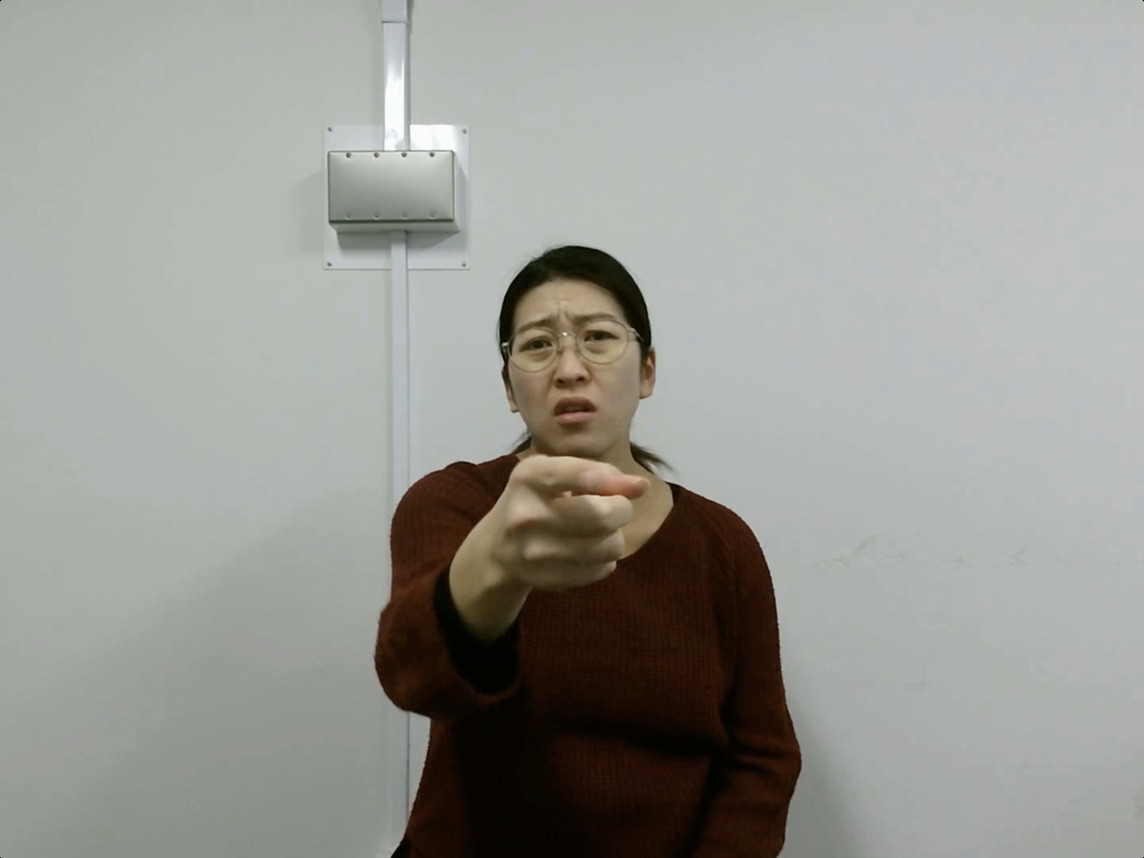} & 
  \includegraphics[width=\figw,trim={4cm 0 4cm 3cm},clip]{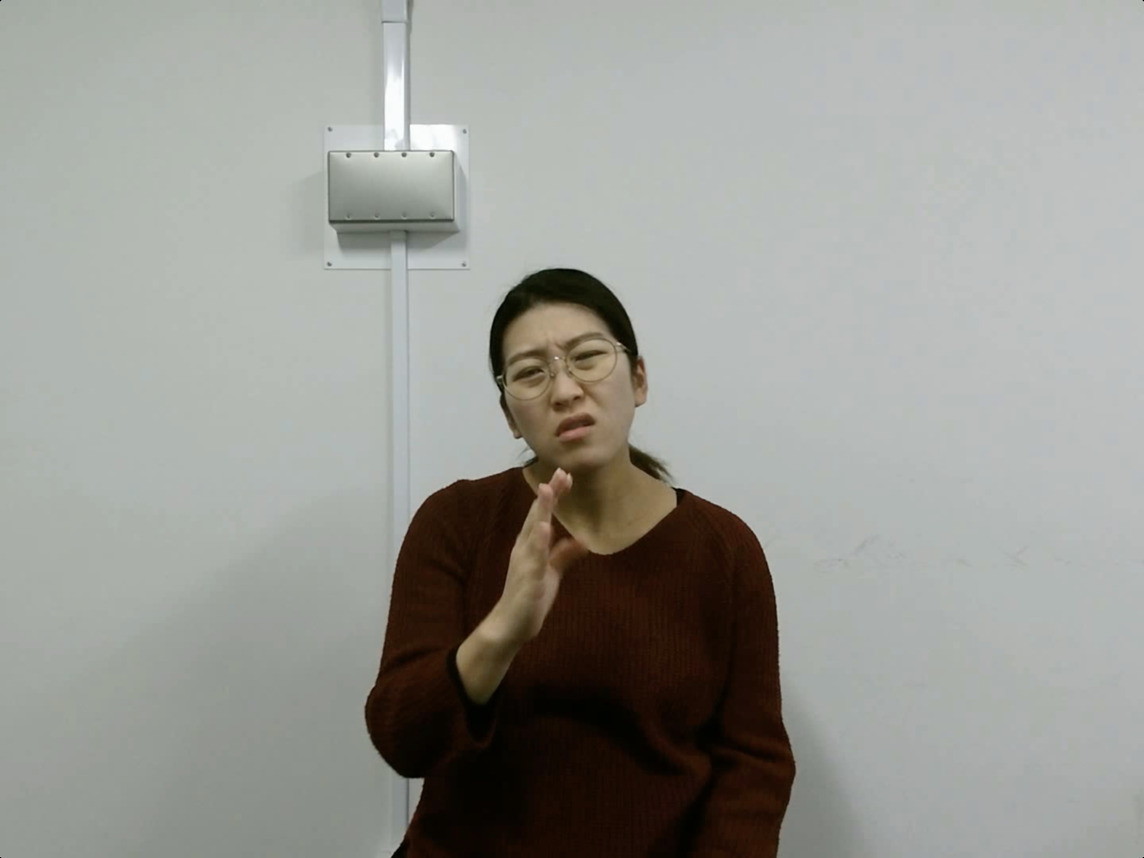} &
  \includegraphics[width=\figw,trim={4cm 0 4cm 3cm},clip]{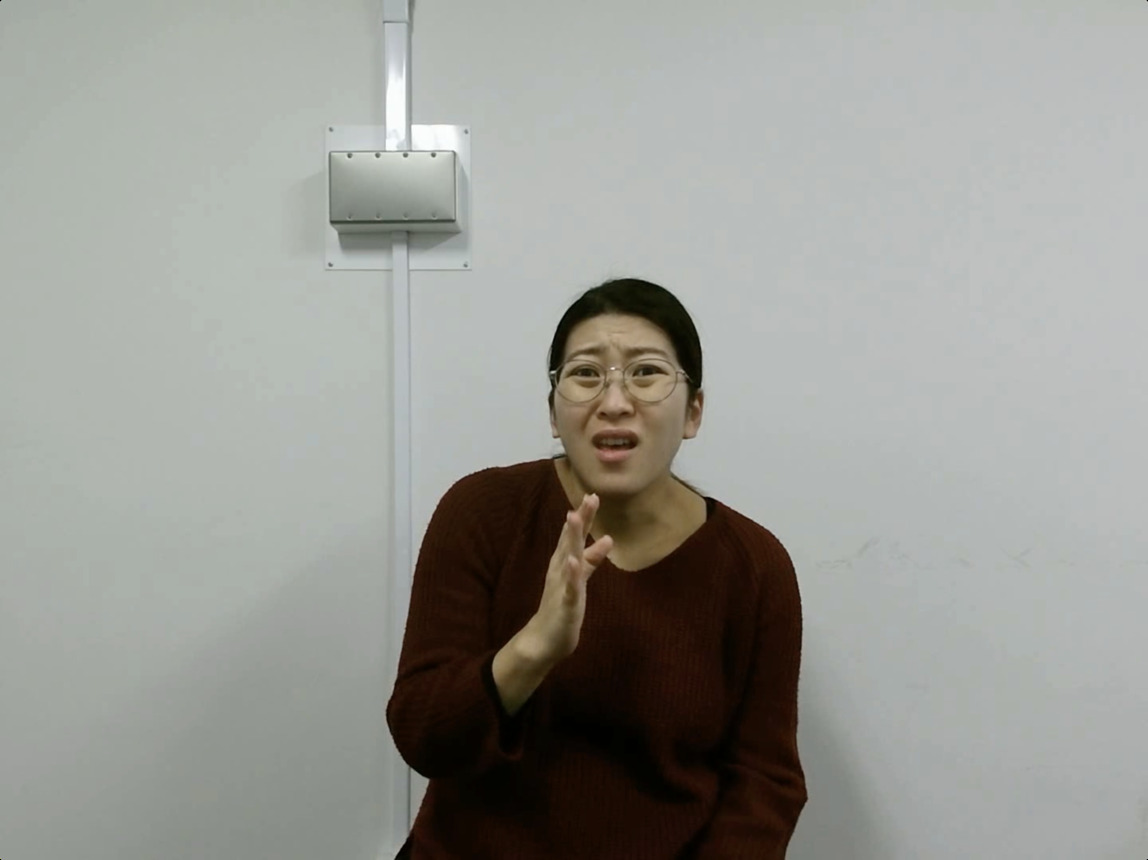} &
  \includegraphics[width=\figw,trim={4cm 0 4cm 3cm},clip]{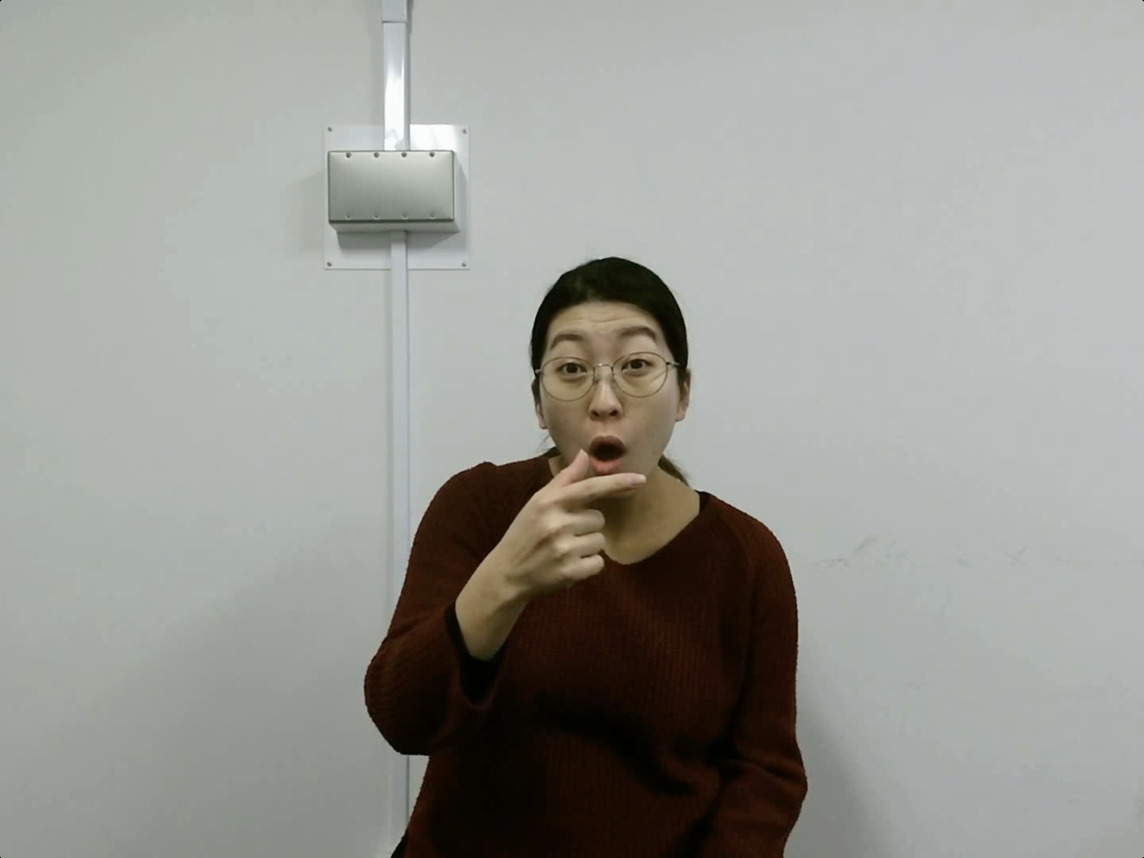} 
  \\
\end{tabular}
}
\caption{Examples of six different emotional expressions for the same utterance \textit{``What? That’s definitely a lie, right? Hurry and say it was a lie.''}  by a signer in the eJSL dataset. 
}\label{fig:emotion_examples}
\end{figure*}

\section{Datasets}
\label{sec:datasets}

We use three sign language datasets: eJSL, EmoSign, and BOBSL.
We use eJSL and EmoSign for evaluation and BOBSL for both evaluation and neural model training. Although all three datasets are in different sign languages, as our focus is para-linguistic (or even non-linguistic) and we are at a very early stage of research, we assume the impact of the  differences is marginal.\footnote{We recognize that this is a strong assumption, and interlingual and intercultural differences will naturally have an impact, but we believe these will only emerge as issues once more research has progressed and recognition performance has improved. \textit{When and how cultural variations limit this assumption} is an important research question for future work.}

\subsection{eJSL}
The {eJSL} (emotional Japanese Sign Language) dataset is our original video corpus containing 78 distinct utterances. As illustrated in Figure~\ref{fig:emotion_examples}, each utterance performed by one male and one female signer across the six Ekman's basic emotions (\textit{anger}, \textit{disgust}, \textit{fear}, \textit{joy}, \textit{sadness}, \textit{surprise})~\cite{Ekman1992} and the neutral state, yielding 1,092 clips in total. 
The signers are native JSL signers who work as vocational deaf actors.
The signers can also read and write fluently in Japanese as well as non-signers. Thus all instructions and utterances were textually presented in Japanese.

Each clip is a complete 
JSL utterance with a single intended emotion. 
The 78 utterances were adopted from a public transcript\footnote{\url{https://github.com/mmorise/ita-corpus/blob/main/emotion_transcript_utf8.txt}}
with substantial modifications in consultation with a professional sign language interpreter so that signers have less difficulties in uttering (e.g., replacing proper names with pronouns, avoiding onomatopoeic words, etc.).

\subsection{EmoSign}
{EmoSign}~\citep{chua2025emosign} is an ASL dataset of 200 clips drawn from the ASLLRP corpus~\citep{neidle2022aslvideocorpora}, designed for affective analysis. We use its \textit{Single Expression Set} of 140 clips, 
which are labeled with a single dominant emotion. 
It covers ten emotion categories (for mapping to our label set, see Appendix~\ref{app:EmoSing2eJSL}) and serves for model comparison, as \citet{chua2025emosign} provide established baselines using vision-capable large language models.

\subsection{BOBSL}\label{sec:BOBSL}
The {BOBSL} dataset~\citep{albanie-etal-2021-bobsl} contains over 1,460 hours of British Sign Language video data from BBC programs by 39 sign language interpreters. 
Using the official subtitle and alignment data from BOBSL, we derive a dataset by applying a textual emotion recognition (TER) model to subtitles, producing large-scale weak labels in seven basic emotions according to the steps below. 

First we extract two base subsets: \textbf{BOBSL-A} from automatically subtitle-alignned data (113,826 clips), 
and \textbf{BOBSL-M} manually subtitle-aligned data (34,046 clips).  

A portion of BOBSL-M is held out (1438 clips) and manually annotated by two English-speaking non-signers based on subtitles, with a high-confidence overlap set (\textbf{BOBSL-M\_C}, 930 clips for testing) showing moderate-to-substantial agreement on emotion labels 
(see Appendix~\ref{app:BOBSL_human-annotation}) . 

Finally, we apply a pre-trained TER model
\footnote{\url{https://huggingface.co/michellejieli/emotion_text_classifier}} 
to 
BOBSL-A, 
as we identified the model works best according to our preliminary verification using {BOBSL-M\_C}.
We refer to the 
resulting emotion-annotated dataset 
as
\textbf{BOBSL-A\_TEA} for training. 

\section{Experiments}

In this section, we validate our three hypotheses: (1) TER on subtitles mitigates the data scarcity issue in sign language, (2) selecting temporal segments less affected by grammatical expressions improves accuracy, and (3) hand gesture features enhance recognition beyond facial features alone. 

\subsection{Emotion recognition models and metrics}\label{sec:models_and_metrics}

Through our experiments, we adopt EMO-AffectNet~\citep{ryumina2022cross} for video-based face emotion recognition (FER), with a minor extension to include hand gesture features.
For our hand gesture extension, see Appendix~\ref{app:EAN}.

\citet{ryumina2022cross} provide a comprehensive cross-corpus study covering eight emotion datasets. 
Their framework combines a ResNet-50 FER backbone, pretrained on VGGFace2, with temporal modeling modules
using multiple data augmentation strategies and label balancing.
As our primary baseline, we use their public model weights\footnote{\url{https://github.com/ElenaRyumina/EMO-AffectNetModel}}, which were optimized in such a way with non-signer data.
Here after we refer to the plain Emo-AffectNet architecture as EAN and the hand gesture extended version as EANwH.

In accordance with the emotion recognition literature, 
we use weighted accuracy (wAcc) and macro F1 as performance metrics.

\subsection{TER-based automatic data labeling}

To validate the effectiveness of TER-based automated labeling on sign language datasets, we fine-tuned the baseline EAN model with the BOBSL-A\_TEA datasets introduced in section \ref{sec:BOBSL}. 

\begin{table}[t]
\centering
\resizebox{\linewidth}{!}{
\begin{tabular}{lrr}
\toprule
\textbf{Method} & wAcc (\%) & macro F1 (\%) \\
\midrule
EAN w/ non-signers data   & 15.54 & 12.12 \\
EAN w/ BOBSL-A\_TEA       & \textbf{27.85} & \textbf{17.75} \\
\bottomrule
\end{tabular}
}
\caption{Performance of fine-tuning with TER-based labeling on BOBSL-M\_C.}\label{tab:H1_bobsl_C}
\vspace{3mm}
\resizebox{\linewidth}{!}{
\tabcolsep=5pt
\begin{tabular}{lrr}
\toprule
\textbf{Method} & wAcc (\%) & macro F1 (\%)  \\
\midrule
EAN w/ non-signers data  & 7.41 & 9.25 \\
EAN w/ BOBSL-A-TEA       & \textbf{15.11} & \textbf{12.11} \\
\bottomrule
\end{tabular}
}
\caption{Performance of fine-tuning with TER-based labeling on eJSL.}
\label{tab:H1_ejsl}
\end{table}

As shown in Table~\ref{tab:H1_bobsl_C} and Table~\ref{tab:H1_ejsl},
finetuning with BOBSL-A-TEA improved recognition performance significantly not only on BOBSL but also on eJSL, although the current overall performance is still quite low in comparison to that on non-signers.
Nevertheless, the results support our hypothesis that TER-based weak labeling would mitigate the scarcity of sign language emotion recognition data.

\subsection{Temporal segment selection}

If GFEs really obscure affective cues, 
selecting non-signing temporal segments used for FER should improve the recognition performance. This has been theoretically expected but has not been verified quantitatively yet.
Especially, by observation, post-signing segments seem to be emotionally salient, at least in acted eJSL.

Therefore, we compare the following three 
strategies:
(1) using full clip, which is equivalent to the previous experiment settings for Table~\ref{tab:H1_bobsl_C} and Table~\ref{tab:H1_ejsl};
(2) randomly selecting a 2-second segment in each clip; and
(3) using the post-signing 2-second segment in each clip.\footnote{We extracted the post-signing segment automatically using motion intensity, for we instructed signers to remain still for 3 seconds after signing each sentence.}

As expected, the results shown in Table~\ref{tab:segmentation_comparision} confirm temporal segment selection of non-signing or emotionally salient segments is quite effective.

\begin{table}[t]
\centering
\resizebox{\linewidth}{!}{
\begin{tabular}{lrr}
\toprule
\textbf{Method} & wAcc (\%) & macro F1 (\%) \\
\midrule
Full Clip Input             & 15.11 & 12.11 \\
Random 2s Segment           & 15.20 & 12.29 \\
Post-Signing 2s Segment        & \textbf{23.17} & \textbf{19.26} \\
\bottomrule
\end{tabular}
}
\caption{Comparison of temporal segment selection strategies on eJSL.}\label{tab:segmentation_comparision}
\end{table}

\subsection{Incorporating hand motion}

Hand motions are expected to serve cues for signing segments. 
Then, by incorporating hand features, a model would learn an effective way to attend only to non-signing moments.
To confirm this possibility, we applied EANwH (see section~\ref{sec:models_and_metrics}), a hand-feature extended version of EAN.

As expected, the results shown in Table~\ref{tab:modality-C} and Table~\ref{tab:modality-ejsl} confirm that incorporating hand features
are effective both for BOBSL and eJSL.
For eJSL, EANwH using full clips performs better than the post-signing segment selection with EAN.

\begin{table}[t]
\centering
\small
\begin{tabular}{lrr}
\toprule
\textbf{Method} & wAcc (\%) & macro F1 (\%) \\
\midrule
EAN (full clip)   & 27.85 & 17.75 \\
EANwH (full clip) & \textbf{32.72} & \textbf{20.03} \\
\bottomrule
\end{tabular}
\caption{Performance of EANwH on BOBSL-M\_C.}
\label{tab:modality-C}
\vspace{3mm}
\small
\begin{tabular}{lrr}
\toprule
\textbf{Method} & wAcc (\%) & macro F1 (\%) \\
\midrule
EAN   (full clip) & 15.11 & 12.11 \\ 
EAN   (post 2s) & 23.17 & 19.26 \\
EANwH (full clip) & \textbf{24.63} & \textbf{21.09} \\
\bottomrule
\end{tabular}
\caption{Performance of EANwH on eJSL.}
\label{tab:modality-ejsl}
\end{table}

\subsection{Comparison to vision-capable LLMs}

Finally, we compare our EANwH model to vision-capable LLMs (Qwen 2.5 and GPT-4o) using EmoSign~\cite{chua2025emosign}. 
We applied the procedure presented in \cite{chua2025emosign} and could reproduce mostly the same results with them.\footnote{Only a few clips were differently classified from the results reported in their confusion matrices. We set the temperature parameter 0.}

The results shown in Table~\ref{tab:model_comparison_EmoSign}
suggest that EANwH is better than the tested LLMs. 
Especially, EANwH has superior performance on Neutral.
Appendix Table~\ref{tab:distribution_merge} shows that Neutral is the majority class in BOBSL-M\_C, as is often the case in real-world data. Therefore, performance on the Neutral class may strongly influence users’ perceived performance in practical applications.

Table~\ref{tab:model_comparison_eJSL} shows the evaluation results on eJSL with the same procedure.
EANwH obtained the best results, consistently with the test on BOBSL-M\_C shown in Table~\ref{tab:model_comparison_EmoSign}.

\begin{table}[t]
\centering
\resizebox{\linewidth}{!}{%
\tabcolsep=3pt
\begin{tabular}{l|rrrrrrr|r}
\toprule
\textbf{Model} & \textbf{Joy} & \textbf{Sad.} & \textbf{Ang.} & \textbf{Dis.} & \textbf{Fear} & \textbf{Sur.} & \textbf{Neu.} &  \textbf{Total} \\
\midrule
Qwen2.5                   & 39.18 & 4.26 &  \textbf{28.57} & 0.00 & 0.00 & 17.65 & 10.17  & 14.26 \\
GPT-4o                     & 38.38 & \textbf{27.27} & 0.00 & \textbf{28.57} & 8.33 & 0.00 & 0.00  & 14.65 \\
EANwH & 30.99 & 16.67 & 26.67 & 8.33 & \textbf{10.53} & 0.00 & \textbf{25.00}  & \textbf{16.88} \\
\bottomrule
\end{tabular}
}
\caption{
Per-class F1 and overall macro F1 scores of vision-capable LLMs and EANwH on EmoSign.
}
\label{tab:model_comparison_EmoSign}
\vspace{3mm}
\resizebox{\linewidth}{!}{%
\tabcolsep=3pt
\begin{tabular}{l|rrrrrrr|r}
\toprule
\textbf{Model} & \textbf{Joy} & \textbf{Sad.} & \textbf{Ang.} & \textbf{Dis.} & \textbf{Fear} & \textbf{Sur.} & \textbf{Neu.} &  \textbf{Total} \\
\midrule
Qwen2.5 & 20.91 & \textbf{11.98} & 2.53 & 12.10 & 9.57 & 1.27
 & 19.84 & 11.17\\
GPT-4o  & 7.38 & 4.64 & \textbf{15.93} & \textbf{23.79} & 8.61 & 11.00 & 6.67 & 11.15 \\
EANwH  & \textbf{35.91} & 10.64 & 15.55 & 14.29 & \textbf{9.65} & \textbf{21.10} & \textbf{40.49} & \textbf{21.09}\\
\bottomrule
\end{tabular}
}
\caption{
Per-class F1 and overall macro F1 scores of vision-capable LLMs and EANwH on eJSL.
}
\label{tab:model_comparison_eJSL}
\end{table}

\subsection{Discussion}
\label{sec:discussion}

Statistical tests on eJSL confirm that EAN w/ BOBSL-A-TEA is significantly better than EAN w/ non-signers (p < 0.0001, Table 2), and EANwH (full clip) is significantly better than EAN (full clip) (p < 0.0001, Table 5).
However, the gain from EAN (post 2s) to EANwH (full clip) is not significant.

While we observed gains from the simplest baseline, 
the achieved overall performance is still very limited.\footnote{
We sampled 70 clips (10 per class) of one signer and asked the other signer to classify them. The achieved macro F1 score was 77.78, while a non-signer achieved 57.85 on the same 70 clips. 
The former would be the upper-bound and the latter would be the lower-bound for practical applications. 
} 
However, as EANwH, our hand motion-enhanced version, is also quite naive, there should be much room for technical improvements.

Utilization of existing resource will also enhance performance.
While this paper utilized annotated data in a cross-lingual setting, use of more datasets from the same sign language in training will improve results.

Fundamentally both EAN and EANwH do not understand signed linguistic content in utterances. As discussed in section~\ref{sec:er_and_sl}, linguistic content can serve strong emotion indicators in usual situations. 
Thus, integration with sign language understanding also must be be explored. 

\section{Conclusion}

To push forward the research on emotion recognition in signers,
this paper introduced a new sign language benchmark dataset eJSL, in which two JSL signers acted seven emotions for 78 utterances.

With eJSL and other two datasets, i.e., BOBSL and EmoSign, we empirically demonstrated 
effectiveness of textual emotion recognition, temporal segment selection and hand motion.
We hope our eJSL and findings contribute emotion recognition in signers and sign language. 

\newpage
\section{Limitations}

As discussed in section~\ref{sec:discussion}, the current achieved performance of emotion recognition in signers is very limited. Therefore, the findings in this paper may not be applicable after the performance is significantly improved in future.

Our eJSL dataset contains only two JSL signers. The data may not be representative of the JSL community. In addition, the data are acted and may be different from real spontaneous data. (Note that, however, the current standard emotion recognition datasets in English~\cite{Busso2008,poria2019meld} are also acted.)

eJSL marks the first step in research into emotion recognition associated with sign languages (especially JSL), and has a certain significance as a benchmark. However, since it only includes two signers, it will be essential to use it in conjunction with other datasets.
Increasing the number of signers of eJSL itself is also an important future challenge.

The dataset may be used for purposes other than benchmarking, such as analysis by linguists, but should not be used to train models without careful consideration.

\section{Ethical Considerations}
The eJSL data collection was conducted in February 2025 after obtaining the signers' consents using the standard consent form of our institute. 
In accordance with the institutional ethics committee's ethical review requirements checklist, the data collection was exempt from full ethical review in advance. 
The interpreter and signers were appropriately compensated in accordance with the institutional regulations.

The eJSL dataset has been reviewed by two signatories prior to its release. It is available for loan on the condition that it is not redistributed and is used for research purposes only.



%
%

\bibliographystyle{apalike}
\bibliography{ref,ref_add}

\appendix
\clearpage

\section{Mapping from EmoSign to Ekman's Basic Emotions}\label{app:EmoSing2eJSL}

We map EmoSign surpris\_ pos and surprise\_neg
to surprise, worry to fear and frustration to sadness, 
based on semantic similarity~\cite{Russell1980}.
Table~\ref{tab:distribution_merge} shows this mapping and the original counts.

\begin{table}[h]
\centering
\begin{tabular}{llr}
\toprule
\textbf{eJSL (Ekman)} &
\textbf{EmoSign} & \textbf{Count} \\
\midrule
Joy
& Happyness       & 54 \\
\midrule Sadness
& Sadness         & 10 \\
& Frustration     & 19 \\
\midrule Anger
& Anger           & 3  \\
\midrule Disgust
& Disgust         & 10 \\
\midrule Fear
& Fear            & 7 \\
& Worry           & 14 \\
\midrule Surprise
& Surprise\_pos   & 5  \\
& Surprise\_neg   & 7 \\
\midrule Neutral
& Neutral         & 11 \\
\bottomrule
\end{tabular}
\caption{Emotion distribution of the EmoSign single expression set (N=140) and mapping to Ekman's basic emotions.}
\label{tab:distribution_merge}
\end{table}

\section{Manual Emotion Annotation on BOBSL subtitles}\label{app:BOBSL_human-annotation}

In the BOBSL-M subset, we manually annotated a selected subset of 1,438 clips for emotion labels using two independent annotators, both of whom labeled the same set of video segments (1 male and 1 female students who graduated universities in North America). Based on these annotations, we created two subsets: \textbf{BOBSL-M\_A1} and \textbf{BOBSL-M\_A2}, corresponding to the individual annotations from each annotator. The intersection of segments where both annotators provided consistent labels forms a high-confidence subset named \textbf{BOBSL-M\_C}
(Table \ref{tab:emotion_count_m-tea}).

\begin{table}[h]
\centering
{
\begin{tabular}{lrrrr}
\toprule
\textbf{Emotion} & \textbf{M\_A1} & \textbf{M\_A2} & \textbf{M\_C} \\
\midrule
Joy      & 59 & 251 & 48 \\
Sadness  & 37 & 110 & 25 \\
Anger    & 35 & 92 & 26 \\
Disgust  & 19 & 55 & 10 \\
Fear     & 21 & 33 & 5 \\
Surprise & 34 & 47 & 8 \\
Neutral  & 1233 & 850 & 808 \\
\midrule
\textbf{Total}   & 1438 & 1438 & 930 \\
\bottomrule
\end{tabular}
}
\caption{Number of instances per emotion of BOBSL-M subsets.}
\label{tab:emotion_count_m-tea}
\end{table}

\paragraph{Annotation Instructions} The annotators were instructed as follows:

\vspace{\baselineskip}
{\tt \small
\noindent
Task description: Use 7 emotion labels to annotate sentences in several text document. Each sentence can only correspond to exactly 1 emotion label. Use the annotation tool to label sentences.\\

\noindent
Emotion category: Anger, Disgust, Fear, Joy, Neutral, Sadness, Surprise. For the “Neutral” label, it is used for the sentence that does not have an obvious emotion. \\

\noindent
How to use the labeling tool: The tool shows the sentence to be annotated and its context, determine the emotion of the sentence to be annotated with its context. When labeling, each emotion maps to a key, just press a key to do the corresponding labeling:
'a': 'Anger', 'd': 'Disgust', 'f': 'Fear', 'j': 'Joy', 'n':'Neutral', 's': 'Sadness', 'u': 'Surprise.\\

\noindent
Examples for each emotion category:\\
** The quoted sentences are from the internet. \\
** The unquoted sentences are from the dataset.\\
\noindent
1. Anger: \\
"I can't believe you did that! How could you be so careless?"\\
What the fuck is wrong with you?! \\
2. Disgust: \\
"The way they treated those poor animals is revolting." \\
Oh, horrible. \\
3. Fear: \\
"I’m really scared about what might happen next. This is terrifying." \\
You must be a nightmare to live with.   \\
4. Joy: \\
"I’m so happy! This is the best news I’ve heard all day!" \\
Everyone was cheering, clapping.   
5. Neutral: \\
"I went to the store today and bought some groceries." \\
It's one of the oldest and grandest houses in Henrietta Street, built in 1743.\\   
6. Sadness: \\
"I’m really feeling down today. Everything seems so hopeless." \\
I regret to inform you, Mr Keys, that Thomas was killed this morning, in Iraq, in the line of duty.   \\
7. Surprise: \\
"Wow, I didn’t see that coming! What a shock!" \\
I just can't believe it, just out there, Daddy, is the inner city London. 
}

\paragraph{Agreement between M\_A1 and M\_A2}
To evaluate the annotation consistency between the two annotators, we computed the {Gwet’s AC1}~\citep{gwet2008computing} as a robust measure of inter-rater agreement. Compared to Cohen’s Kappa~\citep{cohen1960kappa}, AC1 is less sensitive to category imbalance and prevalence issues, making it more appropriate for our dataset, where the {neutral} class dominates the distribution~\citep{feinstein1990high}. The observed agreement ($P_o$) was 0.6467, and the expected agreement by chance ($P_e$) was 0.0762. Using the formula:
\[
\text{AC1} = \frac{P_o - P_e}{1 - P_e}
\]
we obtained a Gwet’s AC1 value of {0.6176}, indicating moderate to substantial agreement between annotators. This level of consistency supports the reliability of the overlapping subset \textbf{BOBSL-M\_C}, which is used as a high-confidence evaluation set in our experiments.

\section{EMO-AffectNet (EAN) and EANwH}\label{app:EAN}

\begin{figure*}[h]
  \centering
  \includegraphics[width=.95\linewidth]{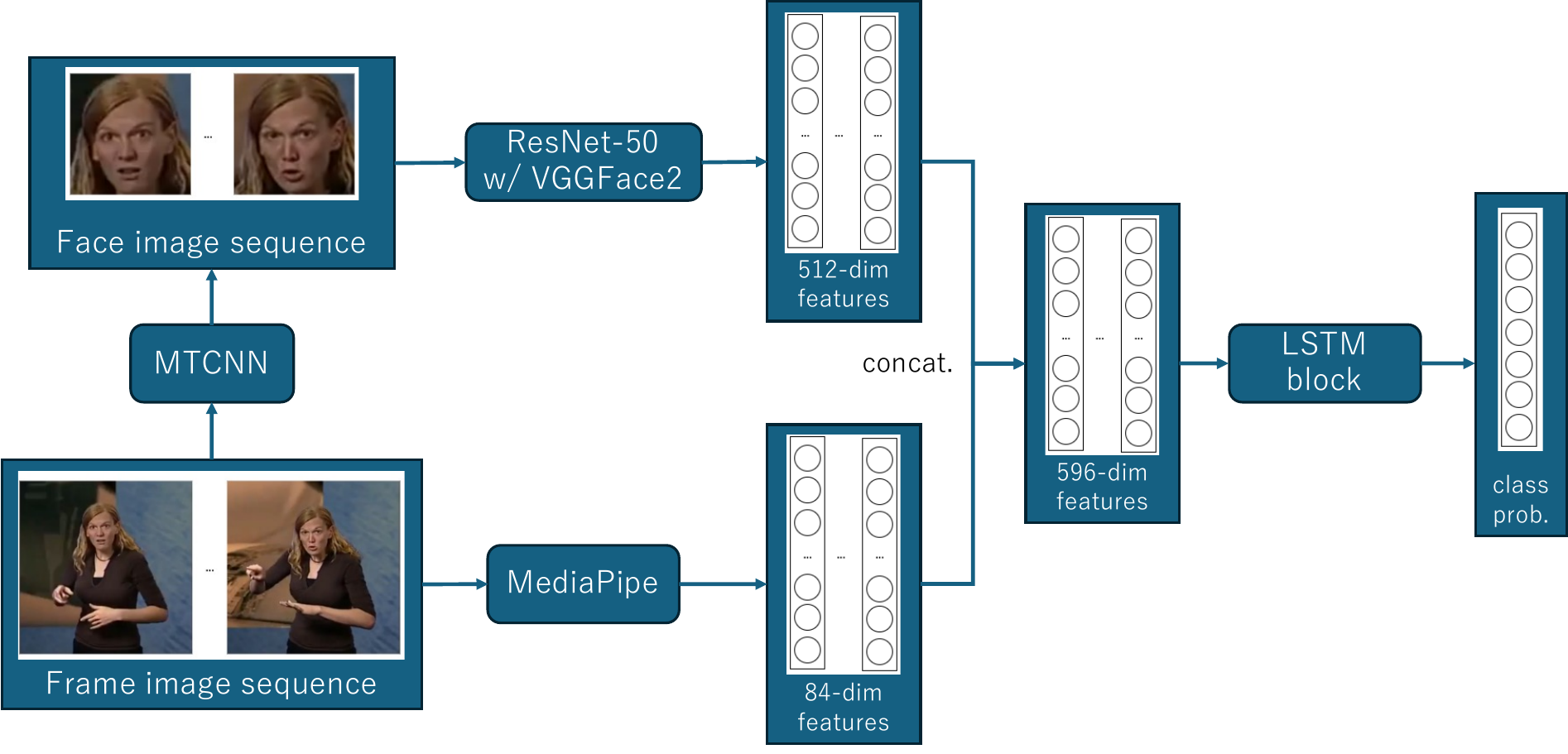}
  \caption{EANwH model architecture using both facial and hand features.}
  \label{fig:face_hand_model}
\end{figure*}

We extend EMO-AffectNet~\cite{ryumina2022cross} to incorporate hand motion as shown Figure~\ref{fig:face_hand_model}.

\subsection{Feature Extraction}

In our model, we extract modality-specific features from both facial images and hand skeletal data. 

\paragraph{Facial Feature Extraction.}
For facial features, we adopt the same methodology as described in the large-scale visual cross-corpus study by \citet{ryumina2022cross}. Specifically, each face image, cropped to $224 \times 224$ resolution using MTCNN, is passed through a ResNet-50 backbone pretrained on VGGFace2. The output is a 512-dimensional embedding extracted from the global average pooling (GAP) layer before the final classification head. This representation captures rich identity-independent emotional features and has been shown to generalize well across datasets with varying demographics and acquisition conditions. The extracted features are stored frame-by-frame as a temporal sequence of fixed-length vectors for downstream sequence modeling.

\paragraph{Hand Feature Extraction.}
For hand features, we follow the approach proposed in the sign language recognition model by \citet{long2024skeletonface}. From each frame, we obtain 21 hand keypoints per hand (totaling 42 keypoints) using the MediaPipe Hands pipeline. These keypoints are represented as 2D coordinates and normalized relative to the wrist joint to ensure translation invariance. We also apply coordinate transformation to align the hand pose into a canonical hand-centered coordinate system, as described in their work. This process effectively reduces spatial variance and emphasizes articulation differences across signs. The final hand representation for each frame is a $42 \times 2$ feature matrix, which is flattened and stored as part of the temporal input sequence.

\subsection{Feature Synchronization and Fusion.}
To effectively integrate facial and hand-derived features, we adopt an early fusion strategy at the frame level. For each frame in the video, the 512-dimensional facial feature vector extracted from the ResNet-50 backbone is concatenated with the flattened 84-dimensional hand skeleton vector (21 keypoints × 2D), resulting in a unified 596-dimensional feature vector. This frame-level concatenation preserves temporal alignment between the two modalities and enables the model to capture low-level interactions between facial expressions and hand gestures.

The sequence of fused multimodal vectors is then passed into a temporal modeling module, which captures the temporal dependencies and emotional dynamics across the video using two LSTM layers
of 512 and 256 hidden units, resulting in about 300M parameters. This early fusion design allows for efficient joint modeling of modality-specific and cross-modal patterns without requiring complex attention-based alignment mechanisms. It also ensures robustness against partial modality noise, as both facial and skeletal information are encoded into a shared temporal embedding space from the outset.

\end{document}